\documentclass[10pt,twocolumn,letterpaper]{article}

\usepackage{iccv}
\usepackage{times}
\usepackage{epsfig}
\usepackage{graphicx}
\usepackage{amsmath}
\usepackage{amssymb}
\usepackage{tikz}
\usepackage{pgfplots}
\usepackage{xspace}
\pgfplotsset{compat=1.13}

\usepackage[breaklinks=true,bookmarks=false]{hyperref}

 \iccvfinalcopy 

\def\iccvPaperID{2958} 

\ificcvfinal\fi
\begin{document}

\title{TAPA-MVS: Textureless-Aware PAtchMatch Multi-View Stereo}

\author{Andrea Romanoni\\
Politecnico di Milano, Italy\\
{\tt\small andrea.romanoni@polimi.it}
\and
Matteo Matteucci\\
Politecnico di Milano, Italy\\
{\tt\small matteo.matteucci@polimi.it}
}

\maketitle

\begin{abstract}
One of the most successful approaches in Multi-View Stereo estimates a depth map and a normal map for each view via PatchMatch-based optimization and fuses them into a consistent 3D points cloud. This approach relies on photo-consistency to evaluate the goodness of a depth estimate. It generally produces very accurate results; however, the reconstructed model often lacks completeness, especially in correspondence of broad untextured areas where the photo-consistency metrics are unreliable.  Assuming the untextured areas piecewise planar, in this paper we generate novel PatchMatch hypotheses so to expand reliable depth estimates in neighboring untextured regions. At the same time, we modify the photo-consistency measure such to favor standard or novel PatchMatch depth hypotheses depending on the textureness of the considered area. We also propose a depth refinement step to filter wrong estimates and to fill the gaps on both the depth maps and normal maps while preserving the discontinuities.  The effectiveness of our new methods has been tested against several state of the art algorithms in the publicly available ETH3D dataset containing a wide variety of high and low-resolution images.
\end{abstract}

\section{Introduction}
Multi-View Stereo (MVS) aims at recovering a dense 3D representation of the scene perceived by a set of calibrated images, for instance, to map cities, to create a digital library of cultural heritage or to help robots navigating an environment.
Thanks to the availability of public datasets \cite{seitz2006comparison,strecha2008,jensen2014large}, several successful MVS algorithms have been proposed in the last decade, and their performance keeps increasing.

Depth map estimation represents one of the fundamental and most challenging steps on which most MVS methods rely. Depth maps are then fused together directly into a point cloud~\cite{zheng14joint,schonberger2016pixelwise}, or into a volumetric representation, such as a voxel grid \cite{savinov2016semantic,blaha2016large} or Delaunay triangulation \cite{labatut2007efficient,vu_et_al_2012,kuhn2017tv,romanoni15a}. In the latter case a 3D mesh is extracted and can be further refined via variational methods \cite{vu_et_al_2012,blaha2017semantically,romanoni2017multi} and eventually labelled with semantics \cite{romanoni2018data}.

Although Machine Learning methods have begun to appear \cite{huang2018deepmvs,wang2018mvdepthnet,yao2018mvsnet}, PatchMatch-based algorithms, emerged some years ago, are still the top performing approaches for efficient and accurate depth map estimation.
The core idea of PatchMatch, pioneered by Barnes \etal \cite{barnes2009patchmatch} and extended for depth estimation by Bleyer \etal \cite{bleyer2011patchmatch}, is to choose for each pixel a random guess of the depth and then propagate the most likely estimates to their neighborhood.
Starting from this idea Sch{\"o}nberger \etal \cite{schonberger2016pixelwise} recently proposed a robust framework able to jointly estimate the depth, the normals, and the pixel-wise camera visibility for each view.

\begin{figure}[t]
\centering
\begin{center}
\setlength{\tabcolsep}{1px}
\begin{tabular}{cc}
\includegraphics[width=0.48\columnwidth]{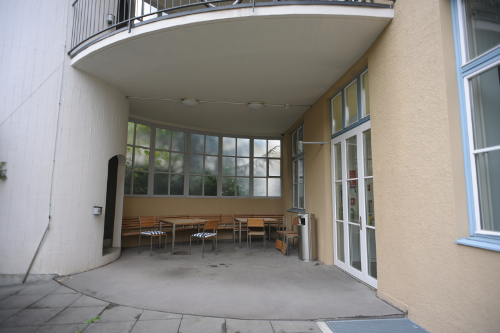}&
\includegraphics[width=0.48\columnwidth]{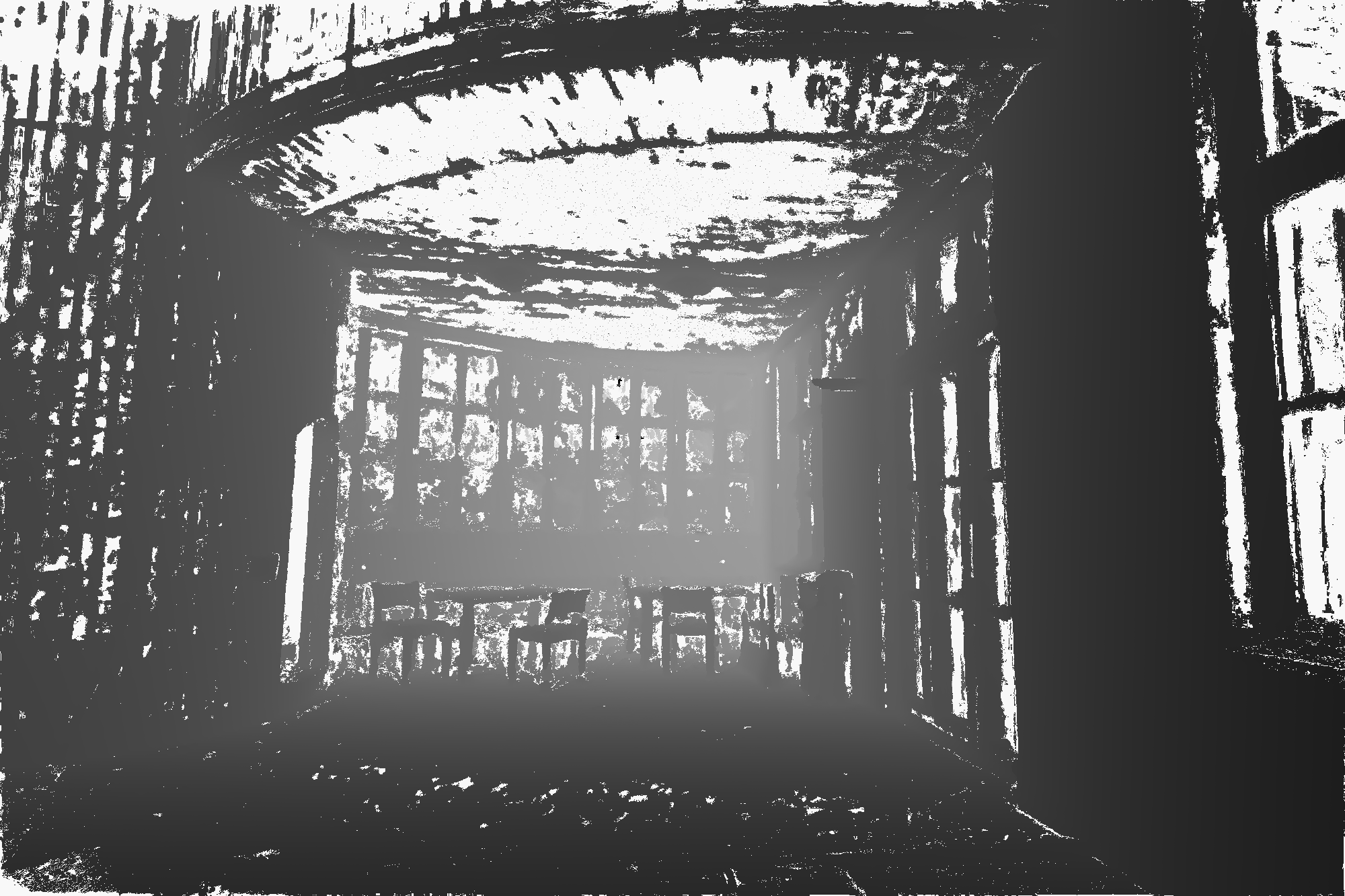}\\
(a) RGB image & (b) COLMAP\\
\includegraphics[width=0.48\columnwidth]{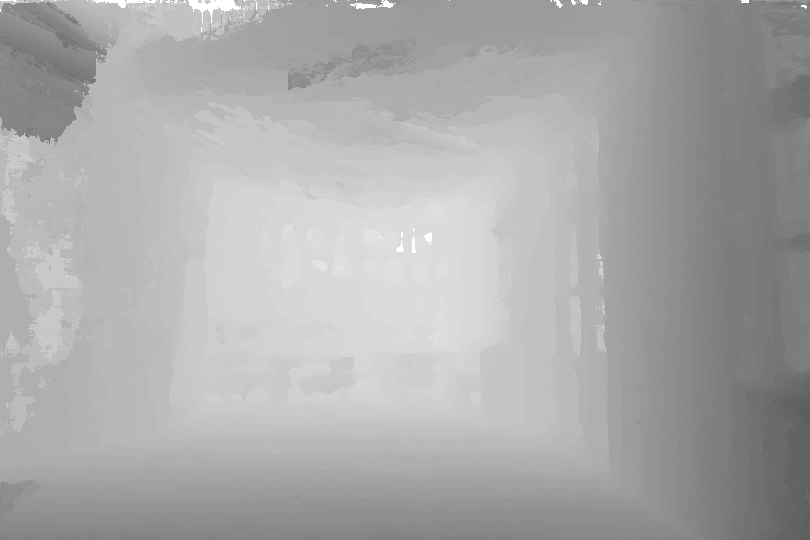}&
\includegraphics[width=0.48\columnwidth]{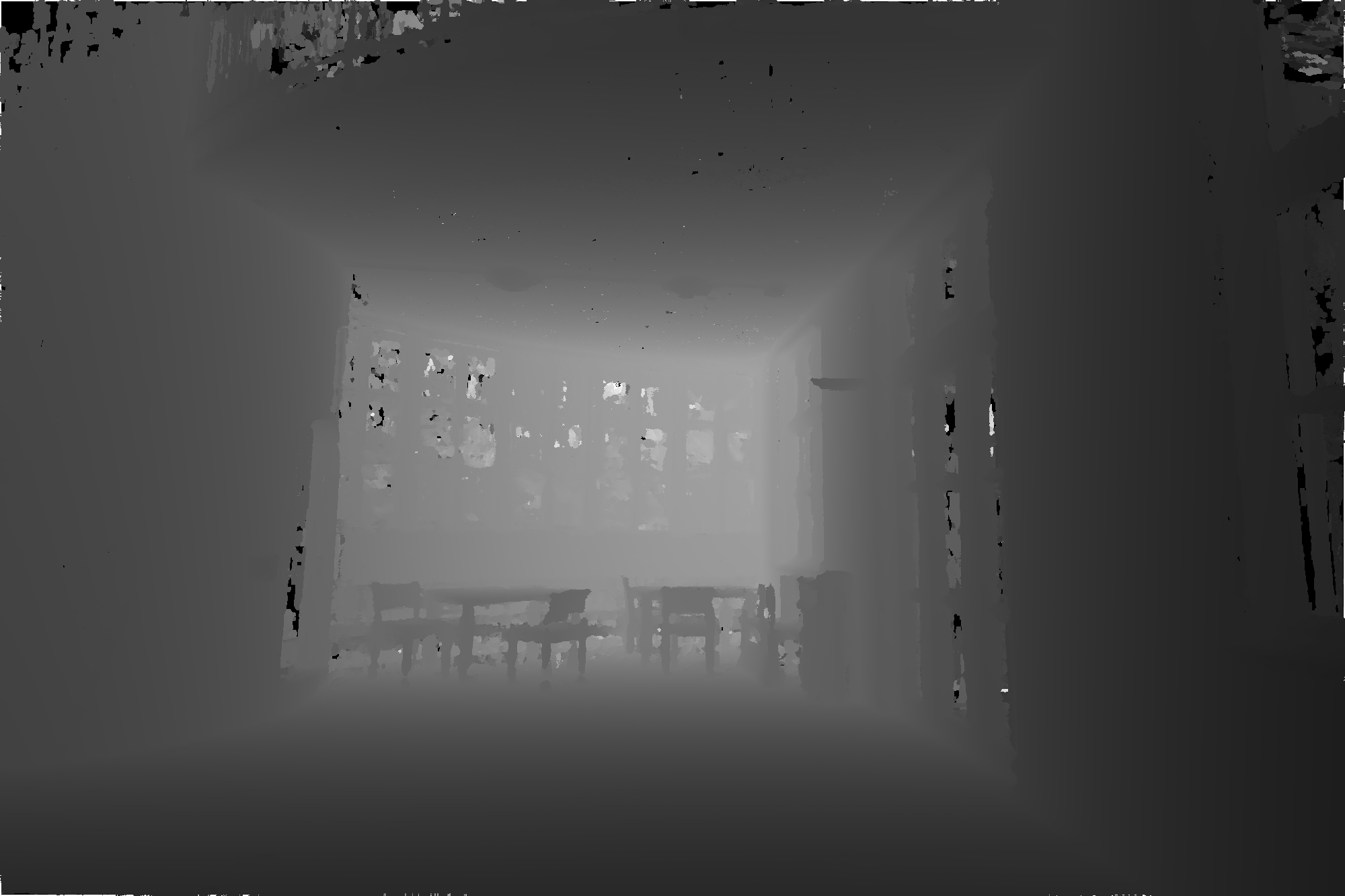}\\
(c) DeepMVS  & (d) TAPA-MVS   \\
\end{tabular}
\caption{Example of the depth map produced by the proposed method with respect to the state-of-the-art}
\label{fig:examples}
\end{center}
\end{figure}

One of the major drawbacks of PatchMatch methods is that most of the untextured regions are not managed correctly  (Figure \ref{fig:examples}(b)). Indeed the optimization highly relies on the photometric measure to discriminate which random estimate is the best guess and to filter out unstable estimates. 
The depth of the untextured regions is hard to be defined with enough confidence since they are homogeneous and thus, the photometric measure alone hardly discerns neighboring regions.

In this paper, we specifically address the untextured regions drawback by leveraging on the assumption that untextured regions are often piecewise flat (Figure \ref{fig:examples}(d)). The framework presented, named TAPA-MVS,  proposes:
\begin{itemize}
    \item a metric to define the textureness of each image pixel; it serves as a proxy to understand how much the photo-consistency metric is reliable.
    \item to subdivide the image into superpixels and, for each iteration of the optimization procedure, to fit one plane for each superpixel; for each pixel, a new depth-normal hypothesis is added and evaluated into the optimization framework considering the likelihood of the plane fitting procedure.
    \item a novel depth refinement method that filters the depth and normal maps and fills each missing estimates with an approximate bilateral weighted median of the neighbors.
\end{itemize}

We tested the proposals against the 38 sequences of the publicly available ETH3D dataset \cite{schops2017multi} (Section \ref{sec:exp}) and the results show that our method is able to significantly improve the completeness of the reconstruction while preserving a very good accuracy. 

In the following after a brief introduction to PatchMatch based methods (Section \ref{sec:patch-match}), we review the COLMAP framework by Sch{\"o}nberger \etal \cite{schonberger2016pixelwise} (Section \ref{sec:review}). Section \ref{sec:method} and \ref{sec:depth} described the proposed texture-aware PatchMatch hypotheses generation and the depth map refinement. Section \ref{sec:exp} illustrates the experimental results.

\section{Patch-Match for Multi-View Stereo}
\label{sec:patch-match}
The PatchMatch seminal paper by Barnes \etal \cite{barnes2009patchmatch} proposed a general method to efficiently compute an approximate nearest neighbor function defining the pixelwise correspondence among patches of two images. 
The idea is to use a collaborative search which exploits local coherency. 
PatchMatch initializes each pixel of an image with a random guess about the location of the nearest neighbor in the second image. 
Then, each pixel propagates its estimate to the neighboring pixels and, among these estimates, the most likely is assigned to the pixel itself. 
As a result the best estimates spread along the entire image.

Bleyer \etal \cite{bleyer2011patchmatch} re-framed this method into the stereo matching realm. Indeed, for each image patch, stereo matching looks in the second image for the corresponding patch, \ie the nearest neighbor in the sense of photometric consistency. 
To improve its robustness the matching function is not limited to fixed sized squared windows, but it extends PatchMatch to estimate a pixel-wise plane orientation adopted to define the matching procedure on slanted support windows.
Heise \etal \cite{heise2013pm} integrated the PatchMatch for stereo into a variational formulation to regularize the estimate with quadratic relaxation. This approach produces smoother depth estimates while preserving edges discontinuities.

The previous works successfully applied the PatchMatch idea to the pair-wise stereo matching problem. The natural extension to Multi-View Stereo was proposed by Shen \cite{shen2013accurate}. Here the author selects a subset of camera pairs depending on the number of shared points computed by Structure from Motion and their mutual parallax angle. Then he estimates a depth map for the selected subset of camera pairs through a simplified version of the method of Bleyer \etal \cite{bleyer2011patchmatch}.
The algorithm refines the depth maps by enforcing consistency among multiple views, and it finally merges the depth maps into a point cloud.

A different multi-view approach by Galliani \etal \cite{Galliani_2015_ICCV} modifies the PatchMatch propagation scheme in such a way that computation can better exploit the parallelization of GPUs. Differently, from Shen \cite{shen2013accurate}, they aggregate, for each reference camera, a set of matching costs compute from different source images.
One of the major drawbacks of these approaches is the decoupled depth estimation and camera pairs selection.
Xu and Tao \cite{xu2018multi} recently proposed an attempt to overcome this issue; they extended \cite{Galliani_2015_ICCV} with a more efficient propagation pattern and, in particular, their optimization procedure jointly considers all the views and all the depth hypotheses.

Rather than considering the whole set of images to compute the matching costs, Zheng \etal \cite{zheng14joint} proposed an elegant method to deal with view selection. They designed a robust method framing the joint depth estimation and pixel-wise view selection problem into a variational approximation framework. Following a generalized Expectation Maximization paradigm, they alternate depth update with a PatchMatch propagation scheme, keeping the view selection fixed, and pixel-wise view inference with the forward-backward algorithm, keeping the depth fixed.

Sch{\"o}nberger \etal \cite{schonberger2016pixelwise} extended this method to jointly estimate per-pixel depths and normals, such that,  differently from \cite{zheng14joint}, the knowledge of the normals enables slanted support windows to avoid the fronto-parallel assumption. Then they add view-dependent priors to select views that more likely induce robust matching cost computation.

The PatchMatch based methods described thus far, have been proven to be among the top performing approachs in several MVS benchmarks \cite{seitz_et_al06,strecha2008,jensen2014large,schoeps2017cvpr}. However, some issues are still open. In particular, most of them strongly rely on photo-consistency measures to discriminate among depth hypotheses. Even if this works remarkably for textured areas and the propagation scheme partially induces smoothness, untextured regions are often poorly reconstructed. 
For this reason, we propose two proxies to improve the reconstruction where untextured areas appear. On the one hand, we seamlessly extend the probabilistic framework to explicitly detect and handle untextured regions by extending the set of PatchMatch hypotheses. On the other side, we complete the depth estimation with a refinement procedure to fill the missing depth estimates.

\section{Review of the COLMAP framework}
\label{sec:review}
In this section we review the state-of-the-art framework proposed by Sch{\"o}nberger \etal \cite{schonberger2016pixelwise} which builds on top of the method presented by Zheng \etal \cite{zheng14joint}.
Let note that in the following, we express the coordinate of the pixel only with a value $l$, since both frameworks sweep independently every single line of the image alternating between rows and columns.

Given a reference image $\mathbf{X} ^ { \mathrm{ ref } }$ and a set of source images $\mathbf{X} ^ { \mathrm{ src } } = \left\{ X ^ { m } | m = 1 \ldots M \right\}$, the framework estimates  the depth $\theta_l$ and the normal $\mathbf{n}_l$ of each pixel $l$, together with a binary variable $Z_l^m\in\{0,1\}$, which indicates if $l$ is visible in image $m$. 
This is framed into a Maximum-A Posteriori (MAP) estimation where the posterior probability is:
{\scriptsize
\begin{multline}
\label{eq:MAP}
P(\mathbf{Z} , \mathbf{ \theta } , \mathbf { N } | \mathbf { X } )  = \frac{ P (\mathbf{Z} , \mathbf{ \theta } , \mathbf { N } , \mathbf { X } )}{P(\mathbf{X})}=\\
=\frac{ 1}{P(\mathbf{X})}
\prod_{l=1}^{L}\prod_{m=1}^{M}\left[P\left(Z_{l,t}^{m}|Z_{l-1,t}^{m},Z_{l,t-1}^{m}\right)\right.\\
\left.P\left(X_{l}^{m}|Z_{l}^{m},\theta_{l},\mathbf{n}_{l},X^{ref}\right)P\left(\theta_{l},\mathbf{n}_{l}|\theta_{l}^{m},\mathbf{n}_{l}^{m}\right)\right],
\end{multline}}
where $L$ is the number of pixels considered in the current line sweep,
$\mathbf { X } = \left\{ \mathbf{X}^{\mathrm{ src }}, \mathrm{X}^{\mathrm{ref}}\right\}$ and $\mathbf{N}=\left\{\mathbf{n}_{l} | l = 1 \ldots L \right\}$.
The likelihood term 
\begin{equation}
\label{eq:like}\scriptsize
P\left(X_{l}^{m}|Z_{l}^{m},\theta_{l}\right)=
\left\{\begin{array}{ll}{\frac{1}{NA} \exp \left(-\frac{\left(1-\rho_{l}^{m}\left(\theta_{l}\right)\right)^{2}}{2\sigma_{\rho}^{2}}\right)} & {\text{if } Z_{l}^{m}=1}\\{\frac{1}{N}\mathcal{U}}&{\text{if } Z_{l}^{m}=0,}\end{array}\right.
\end{equation}
represents the photometric consistency of the patch $X_{l}^{m}$, which belongs to a non-occluded source image $m$ and is around the pixel corresponding to the point at $l$, with respect to the patch $X_{l}^{ref}$ around $l$ in the reference image.
The photometric consistency $\rho$ is computed as a bilaterally weighted NCC, $A = \int_{- 1 } ^ { 1 } \exp \left\{ - \frac { ( 1 - \rho ) ^ { 2 } } { 2 \sigma_{\rho } ^ { 2 } } \right\} d \rho$ and the constant $N$ cancels out in the optimization.
The likelihood term $P\left(\theta_{l},\mathbf{n}_{l}|\theta_{l}^{m},\mathbf{n}_{l}^{m}\right)$ represents the geometric consistency and enforces multi-view depth and normal coherence.
Finally $P\left(Z_{l,t}^{m}|Z_{l-1,t}^{m},Z_{l,t-1}^{m}\right)$ favors image occlusion indicators which are smooth both spatially and along the successive iteration of the optimization procedure.

Being Equation \eqref{eq:MAP} intractable, Zheng \etal \cite{zheng14joint} proposed to use variational inference to approximate the real posterior with a function $q(\mathbf{Z},\mathbf{\theta}, \mathbf{N})$ such that the KL divergence of the two functions is minimized.
Sch{\"o}nberger \etal \cite{schonberger2016pixelwise} factorize  $q(\mathbf{Z},\mathbf{\theta}, \mathbf{N})=q(\mathbf{Z})q(\mathbf{\theta}, \mathbf{N})$ and, to estimate such approximation, they propose a variant of the Generalized Expectation-Maximization algorithm \cite{neal1998view}.
In the E step, the values $(\mathbf{\theta}, \mathbf{N})$ are kept fixed, and, in the resulting Hidden Markov Model, the function $q(Z_{l,t}^m)$ is computed by means of message passing. 
In the M step, viceversa, the values of $Z_{l,t}^m$ are fixed, the function $q(\mathbf{\theta}, \mathbf{N})$ is constrained to the family of Kroneker delta functions $q({\theta_l}, \mathbf{n}_l)=q(\theta_l=\theta_l^*, \mathbf{n}_l^*)$.
The new optimal values of $\mathbf{\theta}_l$ and $\mathbf{N}_l$ are computed as:
\begin{equation}
\scriptsize
\label{eq:optim_depth_norm}
    \left( \hat { \theta }_{l } ^ { \text { opt } } , \hat { \mathbf { n } }_{l } ^ { \mathrm{ opt } } \right) = \underset { \theta_{l } ^ { * } , \mathbf { n }_{l } ^ { * } } { \operatorname { argmin } } \frac { 1 } { | S | } \sum_{m \in S } \left( 1 - \rho_{l } ^ { m } \left( \theta_{l } ^ { * } , \mathbf { n }_{l } ^ { * } \right) \right),
\end{equation}
where $S$ is a subset of sources images, randomly sampled according to a probability $P_l(m)$. Probability  $P_l(m)$ favors images not occluded, and coherent with  three priors which encourage good inter-cameras parallax, similar resolution and camera, front-facing the 3D point defined by ${ \theta }_{ l }^{ * } , \mathbf{n}_{l}^{*}$.

According to the PatchMatch scheme proposed in \cite{schonberger2016pixelwise}, the pair $ \left( \theta_{l } ^ { * } , \mathbf { n }_{l } ^ { * } \right)$ evaluated in Equation \eqref{eq:optim_depth_norm} is chosen among the following set of hypotheses:
{\footnotesize
\begin{multline}
\label{eq:hypotheses}
    \left\{ \left( \theta_{l }, \mathbf{ n }_{l } \right) , \left( \theta_{l - 1}^{\mathrm{ prp } }, \mathbf{ n }_{l - 1 } \right) , \left( \theta_{l}^{\mathrm{ rnd } }, \mathbf{ n }_{l } \right) , \left( \theta_{l }, \mathbf{ n }_{l}^{\mathrm{ rnd } } \right) ,\right. \\
    \left. \left( \theta_{l}^{\mathrm{ rnd } }, \mathbf{ n }_{l}^{\mathrm{ rnd } } \right) , \left( \theta_{l}^{\mathrm{ prt } }, \mathbf{ n }_{l } \right) , \left( \theta_{l }, \mathbf{ n }_{l}^{\mathrm{ prt } } \right) \right\},
\end{multline}}
where $\left( \theta_{l}, \mathbf{ n }_{l} \right)$ comes from the previous iteration,  $\left( \theta_{l-1}, \mathbf{ n }_{l-1} \right)$ is the estimate from the previous pixel of the scan, $\left( \theta_{l}^{\mathrm{ rnd}}, \mathbf{ n }_{l} \right)$ is a random hypothesis and finally, $\theta_{l}^{\mathrm{prt}}$ and $ \mathbf{n}_{l}^{\mathrm{prt}}$ are two small perturbations of the estimates $ \theta_{l}$ and $\mathbf{n}_{l}$.

\begin{figure}[t]
\begin{center}
\setlength{\tabcolsep}{1px}
\begin{tabular}{cc}
\includegraphics[width=0.485\columnwidth,height=0.4\columnwidth]{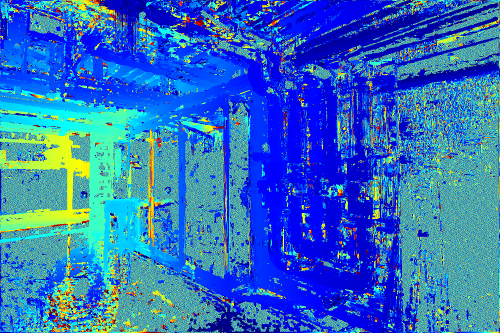}&
\includegraphics[width=0.485\columnwidth,height=0.4\columnwidth]{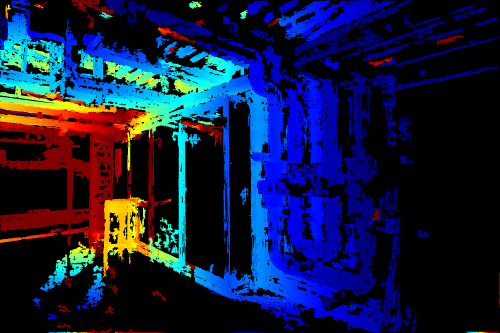}\\
(a)&(b)     
\end{tabular}
\caption{Depth map after the first iteration (a). Unstable regions have been filtered in (b).}
\label{fig:first_iteration}
\end{center}
\end{figure}

\section{Textureness-Aware Joint PatchMatch and View Selection}
\label{sec:method}
The core ingredient that makes a Multi-View Stereo algorithm successful is the quality and the discriminative effectiveness of the stereo comparison among patches belonging to different cameras. Such comparison relies on a photometric measure, computed as Normalized Cross Correlation or similar metrics such as Sum of Squared Differences (SSD), or Bilateral Weighted NCC. 
The major drawback arises in correspondence of untextured regions. Here the discriminative capabilities of NCC become unreliable because all the patches belonging to the untextured area are similar among each other.

Under these assumptions, the idea behind our proposal is to segment images into superpixels such that each superpixel would span a region of the image with a texture mostly homogeneous and it likely stops in correspondence to an image edge. Then, we propagate the depth/normal estimates belonging to photometrically stable regions around the edges to the entire superpixel.
In the following we assume the first iteration of the framework presented in Section \ref{sec:review} is executed so that we have a very first estimation of the depth map, which is reliable only in correspondence of highly textured regions (Figure \ref{fig:first_iteration}).

\subsection{Piecewise Planar Hypotheses generation}
The idea of the method is to augment the set of PatchMatch depth hypotheses in Equation \ref{eq:hypotheses} with novel hypotheses that model a piecewise planar prior corresponding to untextured areas.

In the  first step we extract the superpixels $\mathcal{S} = \{s_1, s_2, \dots, s_{N_{super}} \}$ of  each image by means of the algorithm SEEDS \cite{van2015seeds}.
Since, a superpixel $s_k$ generally contains homogeneous texture, we assume that each pixel covered by a superpixel $s_k$ roughly belongs to the same plane. 

After running the first iteration of depth estimation, we filter out the small  isolated speckles of the depth map obtained (in this paper, with area smaller than $\frac{image area}{5000}$). As a consequence, the area of $s_k$ in the filtered depth map likely contains a set $\mathcal{P}_k^\text{inl}$ of reliable 3D points estimates which roughly corresponds to real 3D points. In the presence of untextured regions, these points mostly belong to the areas near edges (Figure \ref{fig:first_iteration}).

\begin{figure}[t]
\begin{center}
\centering
  \def\svgwidth{0.99\columnwidth}
\begingroup%
  \makeatletter%
  \providecommand\color[2][]{%
    \errmessage{(Inkscape) Color is used for the text in Inkscape, but the package 'color.sty' is not loaded}%
    \renewcommand\color[2][]{}%
  }%
  \providecommand\transparent[1]{%
    \errmessage{(Inkscape) Transparency is used (non-zero) for the text in Inkscape, but the package 'transparent.sty' is not loaded}%
    \renewcommand\transparent[1]{}%
  }%
  \providecommand\rotatebox[2]{#2}%
  \newcommand*\fsize{\dimexpr\f@size pt\relax}%
  \newcommand*\lineheight[1]{\fontsize{\fsize}{#1\fsize}\selectfont}%
  \ifx\svgwidth\undefined%
    \setlength{\unitlength}{577.28504571bp}%
    \ifx\svgscale\undefined%
      \relax%
    \else%
      \setlength{\unitlength}{\unitlength * \real{\svgscale}}%
    \fi%
  \else%
    \setlength{\unitlength}{\svgwidth}%
  \fi%
  \global\let\svgwidth\undefined%
  \global\let\svgscale\undefined%
  \makeatother%
  \begin{picture}(1,0.51030096)%
    \lineheight{1}%
    \setlength\tabcolsep{0pt}%
    \put(0,0){\includegraphics[width=\unitlength,page=1]{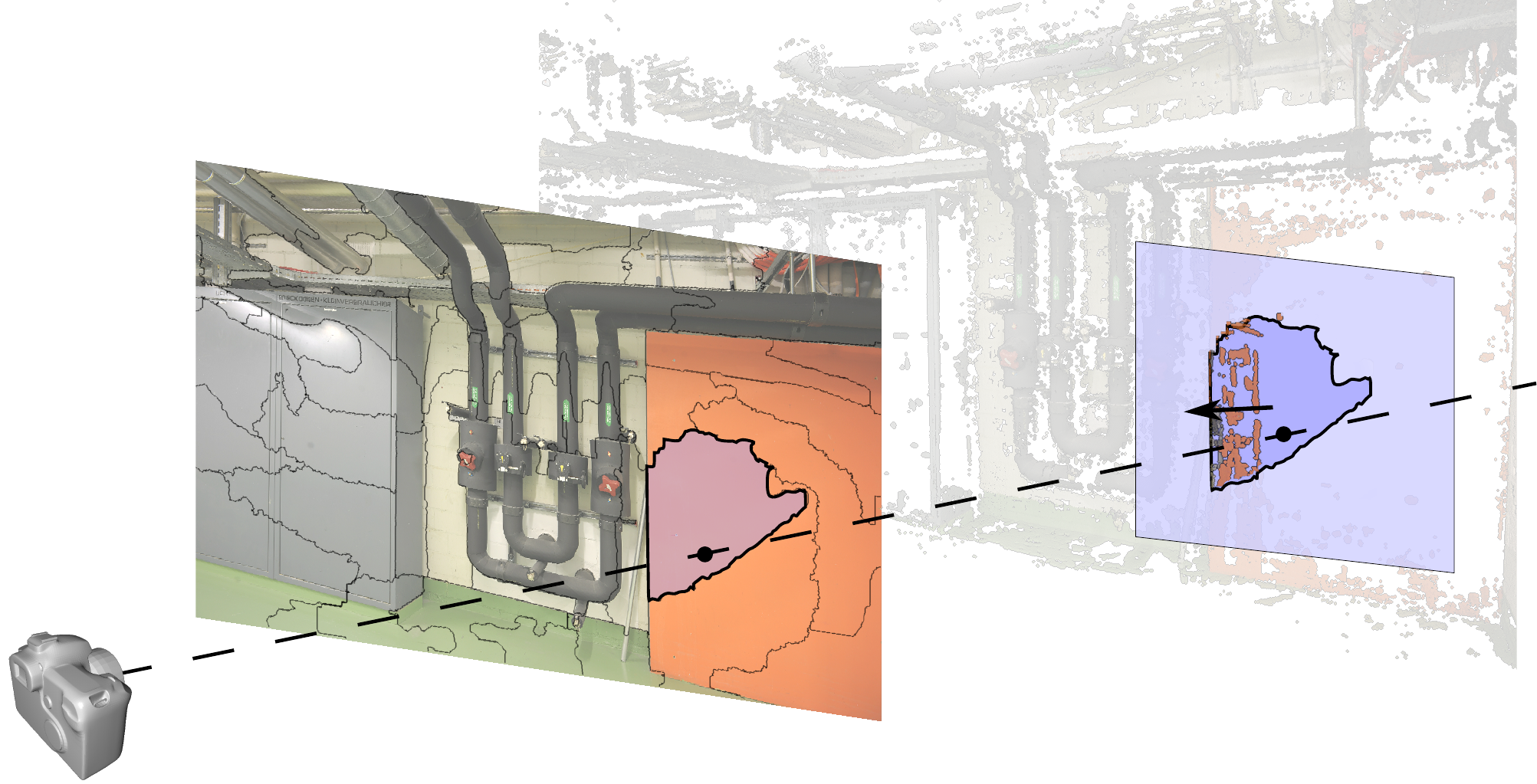}}%
    \put(0.46102067,0.10975835){\color[rgb]{0,0,0}\makebox(0,0)[lt]{\lineheight{1.25}\smash{\begin{tabular}[t]{l}\textbf{$x$}\end{tabular}}}}%
    \put(0.89072061,0.30547502){\color[rgb]{0,0,0}\makebox(0,0)[lt]{\lineheight{1.25}\smash{\begin{tabular}[t]{l}$\pi$\end{tabular}}}}%
    \put(0.75751435,0.25822944){\color[rgb]{0,0,0}\makebox(0,0)[lt]{\lineheight{1.25}\smash{\begin{tabular}[t]{l}\textbf{$n$}\end{tabular}}}}%
    \put(0.83206069,0.18159036){\color[rgb]{0,0,0}\makebox(0,0)[lt]{\lineheight{1.25}\smash{\begin{tabular}[t]{l}\textbf{$\theta_{super}$}\end{tabular}}}}%
  \end{picture}%
\endgroup%

\caption{Depth hypothesis generation. The depth $\theta$ is the distance from the camera to the the plane $\pi$, estimated with the 3D points corresponding to the superpixel extracted on the image. }
\label{fig:hyp}
\end{center}
\end{figure}
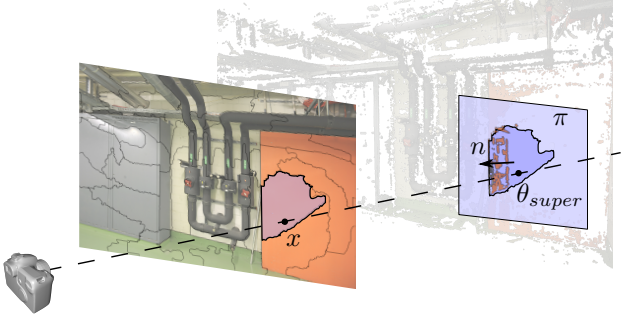

We fit a plane $\pi_k$ on the 3D points in $\mathcal{P}_k^\text{inl}$ with RANSAC, classifying the points farther than 10 cm from the plane as outliers. 
Let us define $\hat{\theta}_x$ the tentative depth hypothesis for a pixel $x$ corresponding to the 3D point on the plane $\pi_k$ and  $\hat{\mathbf{n}}_x$ the corresponding plane normal (Figure \ref{fig:hyp})
Then, let us define the inlier ratio $r_k^{inl} = \frac{\text{num. inliers}}{|\mathcal{P}_k^\text{inl}|}$, whose value expresses the confidence of the plane estimate.

The actual hypotheses $(\theta_x, \mathbf{n}_x)$ for a pixel $x\in s_k$ is generated as follows. 
To deal with fitting uncertainty, we first define $P\left((\theta_x,\mathbf{n}_x) = (\hat{\theta}_x,\hat{\mathbf{n}}_x)\right) = r_k^{inl}$; so that if the value $v_{ran}$ sampled from a uniform distribution is $v_{ran}<=r_k^{inl}$ then $\theta_x = \hat{\theta}_x$.
To propagate the hypotheses from superpixels with good inlier ratio to the neighbors with bad one, if $v_{ran}>r_k^{inl}$ the value of $\theta_x$ is sampled from the neighboring superpixels belonging to a set $\mathcal{N}_k$. 
Since we aim at spreading the depth hypotheses among superpixels with a similar appearance, we sample from $\mathcal{N}_k$ proportionally to the Bhattacharya distance among the RGB histograms of $s_k$ and the elements of $\mathcal{N}_k$.

Experimentally, we noticed that the choice of $N_{super}$, i.e., the number of superpixels, influences how the untextured areas are treated and modeled in our method. 
With small values of $N_{super}$ large areas of the images are nicely covered, but at the same time, limited untextured regions are improperly fused. Vice-versa, a big $N_{super}$ better models small regions while underestimating large areas.
For this reason, we choose to adopt both a coarse and a fine superpixel segmentation of the image such that both small and large untextured areas are modeled properly. 
Therefore, for each pixel, we generate two depth hypotheses: $(\theta_x^{\text{fine}},\mathbf{n}_x^{\text{fine}})$ and $(\theta_x^{\text{coarse}},\mathbf{n}_x^{\text{coarse}})$.
In our experiments we choose $N_{super}^{\text{fine}} = \frac{image width}{20}$ and $N_{super}^{\text{coarse}} = \frac{image width}{30}$.

\begin{figure}[t]
\begin{center}
\setlength{\tabcolsep}{1px}
\begin{tabular}{cc} 
\def\svgwidth{0.49\columnwidth}
\begingroup%
  \makeatletter%
  \providecommand\color[2][]{%
    \errmessage{(Inkscape) Color is used for the text in Inkscape, but the package 'color.sty' is not loaded}%
    \renewcommand\color[2][]{}%
  }%
  \providecommand\transparent[1]{%
    \errmessage{(Inkscape) Transparency is used (non-zero) for the text in Inkscape, but the package 'transparent.sty' is not loaded}%
    \renewcommand\transparent[1]{}%
  }%
  \providecommand\rotatebox[2]{#2}%
  \newcommand*\fsize{\dimexpr\f@size pt\relax}%
  \newcommand*\lineheight[1]{\fontsize{\fsize}{#1\fsize}\selectfont}%
  \ifx\svgwidth\undefined%
    \setlength{\unitlength}{314.21930695bp}%
    \ifx\svgscale\undefined%
      \relax%
    \else%
      \setlength{\unitlength}{\unitlength * \real{\svgscale}}%
    \fi%
  \else%
    \setlength{\unitlength}{\svgwidth}%
  \fi%
  \global\let\svgwidth\undefined%
  \global\let\svgscale\undefined%
  \makeatother%
  \begin{picture}(1,0.99059027)%
    \lineheight{1}%
    \setlength\tabcolsep{0pt}%
    \put(0,0){\includegraphics[width=\unitlength,page=1]{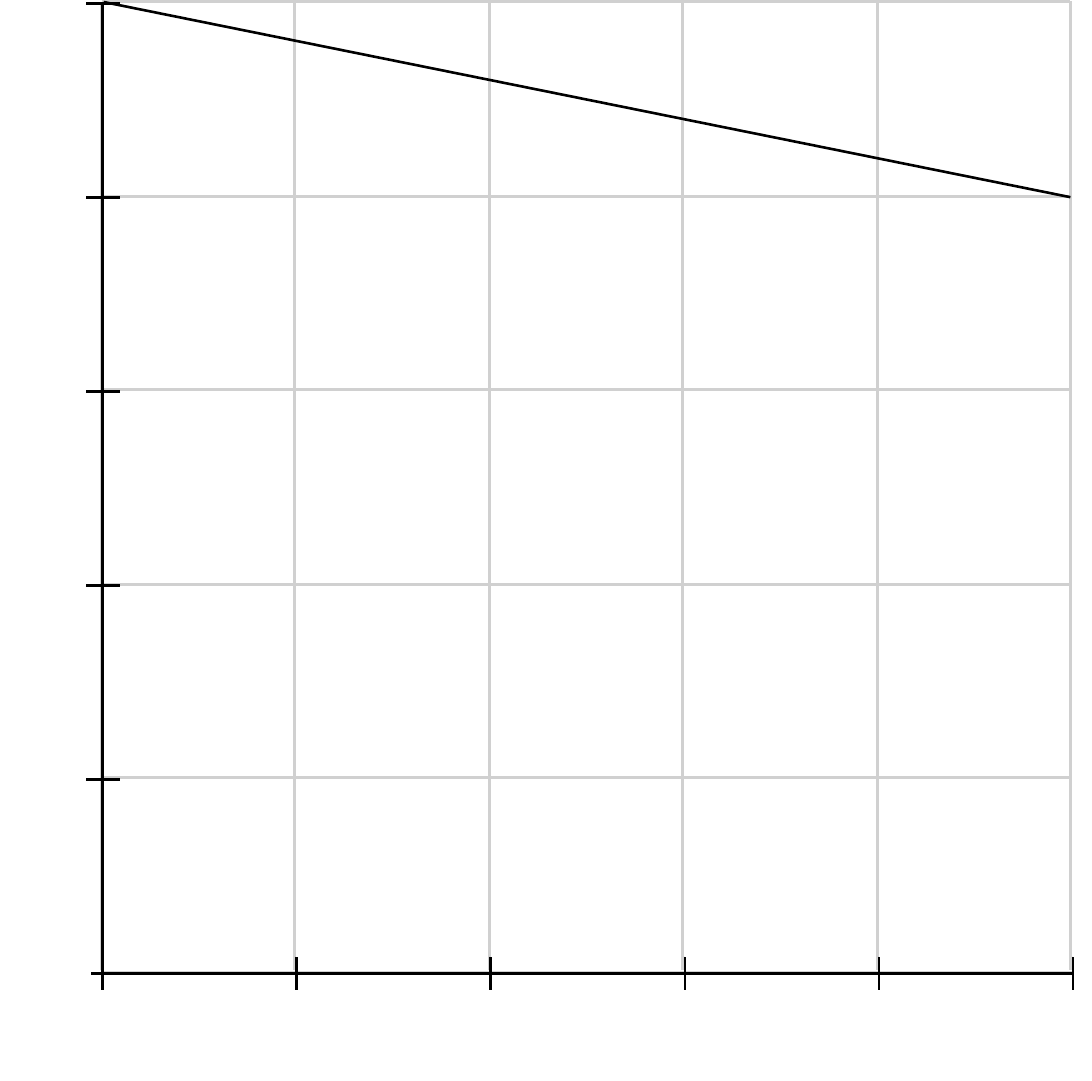}}%
    \put(0.10255077,0.02254916){\color[rgb]{0,0,0}\makebox(0,0)[lt]{\lineheight{0}\smash{\begin{tabular}[t]{l}0\end{tabular}}}}%
    \put(0.04138667,0.97192087){\color[rgb]{0,0,0}\makebox(0,0)[lt]{\lineheight{0}\smash{\begin{tabular}[t]{l}1\end{tabular}}}}%
    \put(0.05401562,0.10994227){\color[rgb]{0,0,0}\makebox(0,0)[t]{\lineheight{0}\smash{\begin{tabular}[t]{c}0\end{tabular}}}}%
    \put(0.98722878,0.02186539){\color[rgb]{0,0,0}\makebox(0,0)[t]{\lineheight{0}\smash{\begin{tabular}[t]{c}1\end{tabular}}}}%
    \put(0.48683927,0.00825614){\color[rgb]{0,0,0}\makebox(0,0)[lt]{\lineheight{1.25}\smash{\begin{tabular}[t]{l}$t_x$\end{tabular}}}}%
    \put(0.02659742,0.47978964){\color[rgb]{0,0,0}\rotatebox{90}{\makebox(0,0)[lt]{\lineheight{1.25}\smash{\begin{tabular}[t]{l}$w_+$\end{tabular}}}}}%
  \end{picture}%
\endgroup%
&
   \def\svgwidth{0.49\columnwidth}
\begingroup%
  \makeatletter%
  \providecommand\color[2][]{%
    \errmessage{(Inkscape) Color is used for the text in Inkscape, but the package 'color.sty' is not loaded}%
    \renewcommand\color[2][]{}%
  }%
  \providecommand\transparent[1]{%
    \errmessage{(Inkscape) Transparency is used (non-zero) for the text in Inkscape, but the package 'transparent.sty' is not loaded}%
    \renewcommand\transparent[1]{}%
  }%
  \providecommand\rotatebox[2]{#2}%
  \newcommand*\fsize{\dimexpr\f@size pt\relax}%
  \newcommand*\lineheight[1]{\fontsize{\fsize}{#1\fsize}\selectfont}%
  \ifx\svgwidth\undefined%
    \setlength{\unitlength}{314.21930695bp}%
    \ifx\svgscale\undefined%
      \relax%
    \else%
      \setlength{\unitlength}{\unitlength * \real{\svgscale}}%
    \fi%
  \else%
    \setlength{\unitlength}{\svgwidth}%
  \fi%
  \global\let\svgwidth\undefined%
  \global\let\svgscale\undefined%
  \makeatother%
  \begin{picture}(1,0.99059027)%
    \lineheight{1}%
    \setlength\tabcolsep{0pt}%
    \put(0,0){\includegraphics[width=\unitlength,page=1]{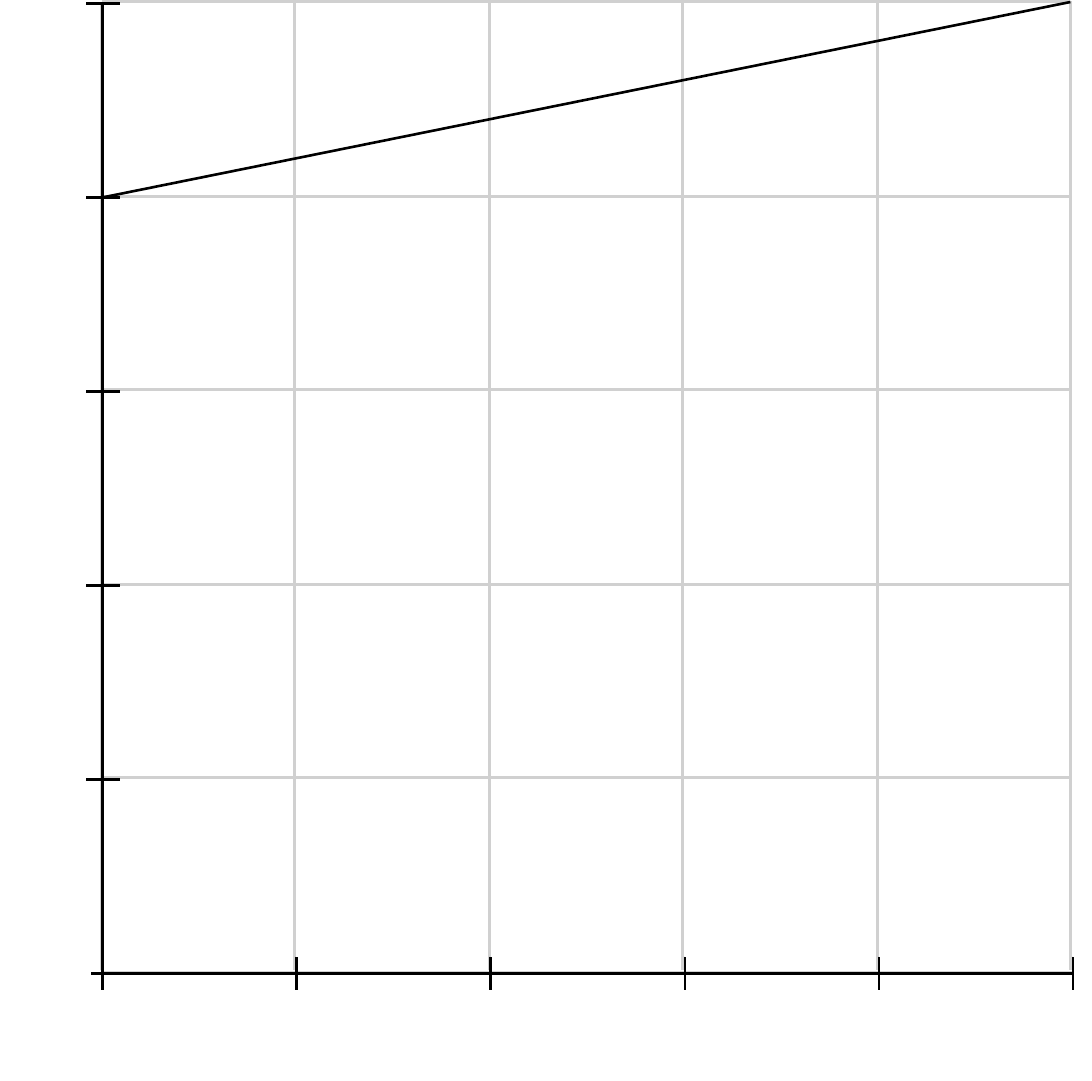}}%
    \put(0.10255077,0.02254916){\color[rgb]{0,0,0}\makebox(0,0)[lt]{\lineheight{0}\smash{\begin{tabular}[t]{l}0\end{tabular}}}}%
    \put(0.04138667,0.97192087){\color[rgb]{0,0,0}\makebox(0,0)[lt]{\lineheight{0}\smash{\begin{tabular}[t]{l}1\end{tabular}}}}%
    \put(0.05401562,0.10994227){\color[rgb]{0,0,0}\makebox(0,0)[t]{\lineheight{0}\smash{\begin{tabular}[t]{c}0\end{tabular}}}}%
    \put(0.98722878,0.02186539){\color[rgb]{0,0,0}\makebox(0,0)[t]{\lineheight{0}\smash{\begin{tabular}[t]{c}1\end{tabular}}}}%
    \put(0.48683927,0.00825614){\color[rgb]{0,0,0}\makebox(0,0)[lt]{\lineheight{1.25}\smash{\begin{tabular}[t]{l}$t_x$\end{tabular}}}}%
    \put(0.02659742,0.47978964){\color[rgb]{0,0,0}\rotatebox{90}{\makebox(0,0)[lt]{\lineheight{1.25}\smash{\begin{tabular}[t]{l}$w_-$\end{tabular}}}}}%
  \end{picture}%
\endgroup%
\\
(a)&(b)     
\end{tabular}
\caption{Weights adopted to tune the photo-consistency and the geometric cost according to the textureness $t_x$}
\label{fig:textureness_w}
\end{center}
\end{figure}
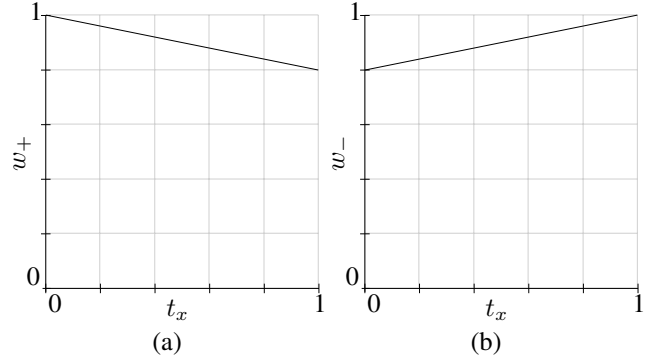

\subsection{Textureness-Aware Hypotheses Integration}
\label{sub:hypo_intr}
To integrate the novel hypotheses into the estimation framework, it is possible to simply add $(\theta_x^{\text{fine}},\mathbf{n}_x^{\text{fine}})$ and $(\theta_x^{\text{coarse}},\mathbf{n}_x^{\text{coarse}})$ to the set of hypotheses defined in Equation \ref{eq:hypotheses}. 
However, in this case, these hypotheses would be treated with no particular attention to untextured areas. Indeed, the optimization framework would compare them against the baseline hypotheses relying on the photo-consistency metric; in the presence of flat evenly colored surfaces, the unreliability of the metric would still affect the estimation process. Instead, the goal of the proposed method is to favor $(\theta_x^{\text{fine}},\mathbf{n}_x^{\text{fine}})$ and $(\theta_x^{\text{coarse}},\mathbf{n}_x^{\text{coarse}})$ where the image presents untextured areas, so to guide the optimization to choose them instead of other guesses.

\begin{figure}[t]
\begin{center}
\setlength{\tabcolsep}{1px}
\begin{tabular}{cc}
\includegraphics[width=0.485\columnwidth,height=0.4\columnwidth]{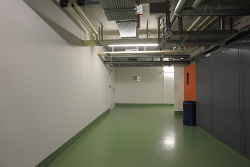}&
\includegraphics[width=0.485\columnwidth,height=0.4\columnwidth]{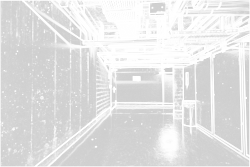}\\
(a)&(b)     
\end{tabular}
\caption{Visualization of the textureness coefficients computed on image (a)}
\label{fig:textureness}
\end{center}
\end{figure}

For these reasons, we first define a pixel-wise textureness coefficient to measure the amount of texture that surrounds a pixel $x$. With a formulation similar to those presented in \cite{vu_et_al_2012}, we define it as:
\begin{equation}
    t_x = \frac{Var_x+\epsilon_{var}}{Var_x+\frac{\epsilon_{var}}{t_{min}}}
\end{equation}
where $Var_x$ is the variance of the 5x5 patch around pixel $x$, $\epsilon_{var}$ is a constant we fixed experimentally at 0.00005, i.e., two order of magnitude smaller than the average variance we found in the ETH3D training dataset (Section \ref{sec:exp}), finally,  $t_{min}=0.5$ is the minimum value we choose for the textureness coefficient; the higher the variance, the closer the coefficient is to 1.0.
Figure \ref{fig:textureness} shows an example of a textureness coefficients image.

To seamlessly integrate the novel hypotheses we use the textureness coefficient to reweight the photometric-based cost $C_{photo} = 1-\rho(\theta,\mathbf{n})$ (Equation \ref{eq:optim_depth_norm}).
Given a pixel $x$ let define two weights:
\begin{equation}
    w^+(x) = 0.8 + 0.2\cdot t_x;
\end{equation}
\begin{equation}
    w^-(x) = 1.0 - 0.2\cdot t_x.
\end{equation}
We use the metric $\bar{C_{photo}} = w^- \cdot C_{photo}$ for the hypotheses contained in the set of Equation \ref{eq:hypotheses} and $\bar{C_{photo}} = w^+\cdot C_{photo}$ for $(\theta_x^{\text{fine}},\mathbf{n}_x^{\text{fine}})$ and $(\theta_x^{\text{coarse}},\mathbf{n}_x^{\text{coarse}})$ so that regions with low texture favors novel hypotheses.
Vice-versa,  it is better to force a higher geometric consistency $C_{geom}$ when we are dealing with the novel hypothesis in the presence of untextured areas. So to keep the formulation simple we use $w^+$ and $w^-$ again turning 
$\bar{C_{geom}} = w^+\cdot C_{geom}$ for the standard set of hypotheses and $\bar{C_{geom}} = w^-\cdot C_{geom}$ for the proposed ones.

\section{Joint Depth and Normal Depth Refinement}
\label{sec:depth}
The hypotheses proposed in the previous section improve the framework estimate accuracy and completeness in correspondence of untextured regions. However, two issues remain open. First, the filtering scheme adopted in \cite{schonberger2016pixelwise} filters out all the estimates that are not photometrically and geometrically consistent among the views. 
Due to their photometric instability, the photo-consistency check removes most of the new depth estimates corresponding to unfiltered areas; therefore, in our case, we neglect this filtering step.

\begin{figure*}[tb]
\centering
\begin{center}
\setlength{\tabcolsep}{1px}
\begin{tabular}{ccc}
\includegraphics[width=0.32\textwidth]{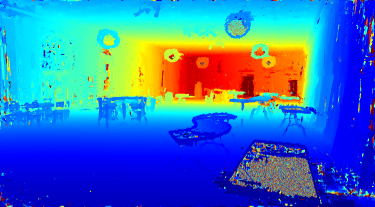}&
\includegraphics[width=0.32\textwidth]{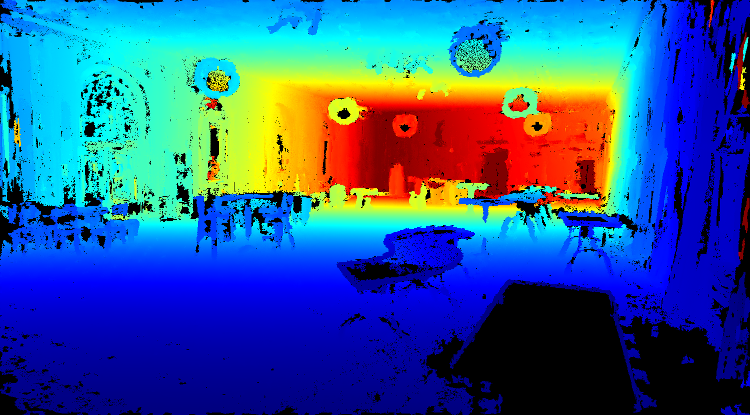}&
\includegraphics[width=0.32\textwidth]{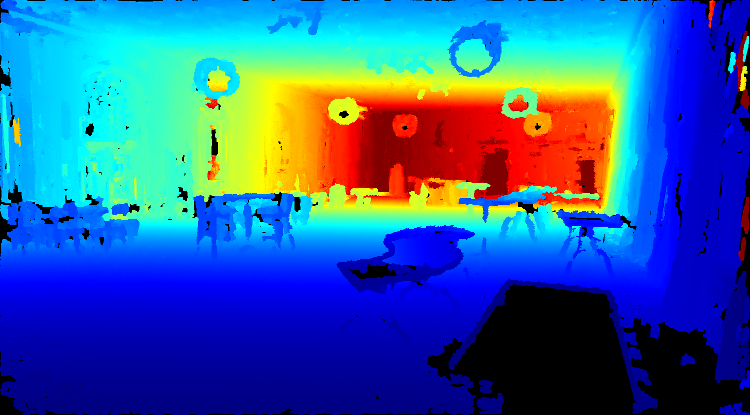}\\
Original Depth Map(a)& After Speckles Removal(b)& After Depth Refinement(c)     
\end{tabular}
\caption{Depth map refinement}
\label{fig:refinement}
\end{center}
\end{figure*}

This leads us to the second issue. The resulting depth map contains wrong and noisy estimates sparsely spread along the depth image (Figure \ref{fig:refinement}(a)). For this reason, we complemented the estimation process with a depth refinement step.
To get rid of wrong estimates that have not converged to a stable solution, we first apply a classical speckles filter to remove small blobs containing non-continuous depths values. We fixed, experimentally, the maximum speckle size of continuous pixels to  $\frac{image area}{5000}$. We consider two pixels as continuous when the depth difference is at most $10\%$ of the scene size.

The output of the filtering procedure contains now small regions where the depth and normal estimates are missing (Figure \ref{fig:refinement}(b)). 
To recover them, we designed the following refinement step.
Let $x_{\text{miss}}$ be a pixel where depth and normal estimates are missing and $\mathcal{N}_{\text{miss}}$ the set of neighboring pixels. 
The simplest solution is to fill the missing estimate by averaging the depth and normal values contained in $\mathcal{N}_{\text{miss}}$. 
A better choice would be to weight the contribution to the average with the bilateral coefficients adopted in the bilateral NCC computation; they give more importance to the pixels close to  $x_{\text{miss}}$ both in the image and in color space.

To better deal with depth discontinuities, we can improve even further the refinement process by using a weighted median of depth and normal instead of the weighted average.
The pixel-wise median and, in particular, the weighted median is computational demanding, thus, to approximate the median computation, we populate a three bins histogram with the depths of the pixels in $\mathcal{N}_{\text{miss}}$. We choose the bin with the highest frequency so to get rid of the outliers, and we compute a bilaterally weighted average of the depth and normals that populates this bin (Figure \ref{fig:refinement}(c)).
The computed depth/normal values are assigned to $x_{\text{miss}}$.

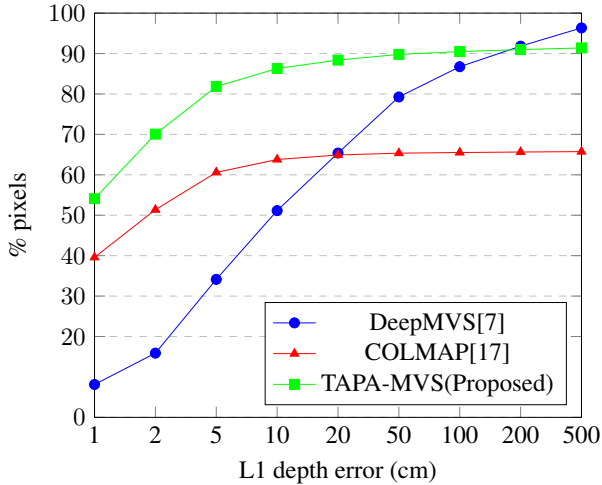
\begin{figure}[tp]
\centering
\resizebox {0.99\columnwidth} {!} {
  \begin{tikzpicture}
  \begin{axis}[
      enlargelimits=false,
      xlabel={L1 depth error (cm)},
      ylabel={\% pixels },
       xmin=1, xmax=500,
       ymin=0, ymax=100,
       symbolic x coords={1,2,5,10,20,50,100,200,500},
      xtick={1,2,5,10,20,50,100,200,500},
      ytick={0,20,30,40,50,60,70,80,90,100},
      legend pos=south east,
      ymajorgrids=true,
      grid style=dashed,
  ]
  \addplot[
      color=blue,
      mark=oplus*]
  table[x index=0,y index=1,col sep=comma]
  {data/depthCompl.txt};
  \addplot[
      color=red,
      mark=triangle*]
  table[x index=0,y index=2,col sep=comma]
  {data/depthCompl.txt};
  \addplot[
      color=green,
      mark=square*]
  table[x index=0,y index=3,col sep=comma]
  {data/depthCompl.txt};
  \legend{DeepMVS\cite{huang2018deepmvs},COLMAP\cite{schonberger2016pixelwise},TAPA-MVS(Proposed)}
  \end{axis}
  \end{tikzpicture}
}
\caption{Depth map error distribution}
\label{fig:depthRes}
\end{figure}

\section{Experiments}
\label{sec:exp}
We tested the proposed method on an Intel(R) Xeon(R) CPU E5-2687W with a GeForce GTX 1080 against the publicly available ETH3D dataset \cite{schops2017multi}.
The dataset is split into test/training and low-/high-resolution for a total of 35 sequences. Parameter tuning is only permitted with the training sequences that are available with the ground truth.
The comparison is carried out by computing the distance from the 3D model to the ground-truth (GT) 3D scans and vice-versa; then, accuracy, completeness, and F1-score are computed considering the percentage of model-to-GT distances below a fixed threshold $\tau$. For a complete description of the evaluation procedure, we refer the reader to \cite{schops2017multi}.
To generate the 3D model out of the depth map estimated with the proposed method, we adopted the depth filtering and fusion implemented in COLMAP. Since depth estimate corresponding to untextured regions can get noisy, we changed the default fusion parameter such that the reprojection error permitted, is more strict (half for high-resolution sequences a quarter for low-resolution ones). On the other hand, even the normal estimate could be noisy, but, usually, the corresponding depths are reasonable. For this reason, we allow for larger normal errors (double the normal error permitted by COLMAP) and demand the outlier filtering to the reprojection error check.

\begin{figure*}[tpb]
\begin{center}
\setlength{\tabcolsep}{1px}
\begin{tabular}{ccccc}
\multicolumn{5}{c}{terrains}\\
\includegraphics[width=0.2\textwidth]{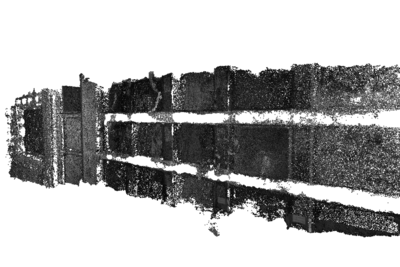}&
\includegraphics[width=0.2\textwidth]{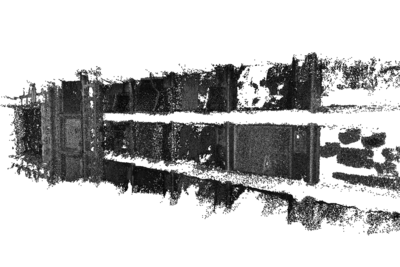}&
\includegraphics[width=0.2\textwidth]{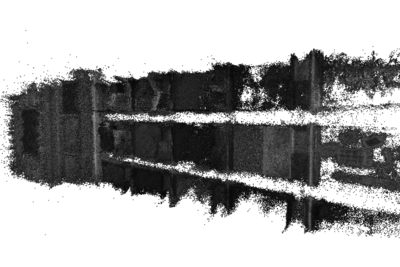}&
\includegraphics[width=0.2\textwidth]{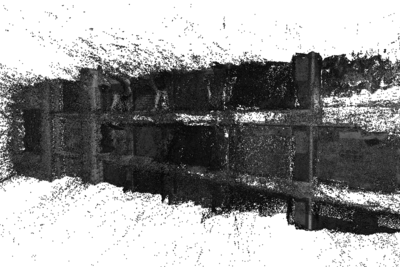}&
\includegraphics[width=0.2\textwidth]{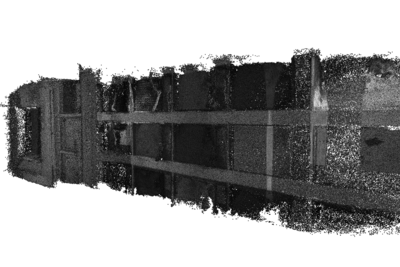}\\
\multicolumn{5}{c}{terrace 2}\\
\includegraphics[width=0.2\textwidth]{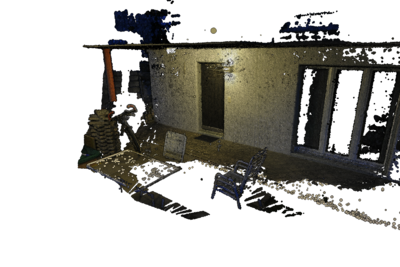}&
\includegraphics[width=0.2\textwidth]{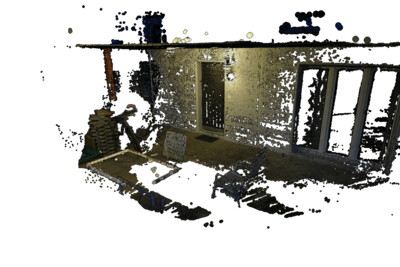}&
\includegraphics[width=0.2\textwidth]{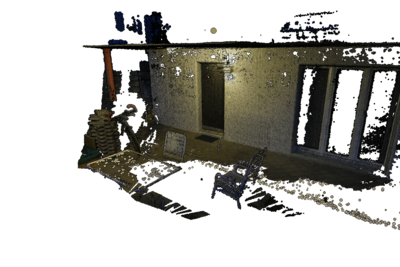}&
\includegraphics[width=0.2\textwidth]{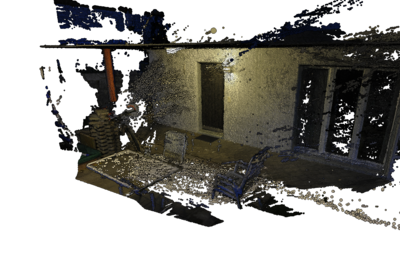}&
\includegraphics[width=0.2\textwidth]{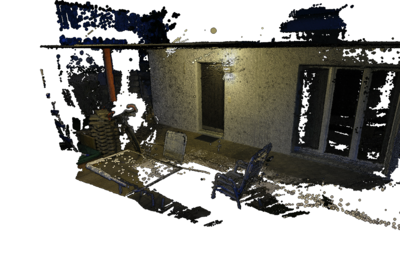}\\
\multicolumn{5}{c}{storage}\\
\includegraphics[width=0.2\textwidth]{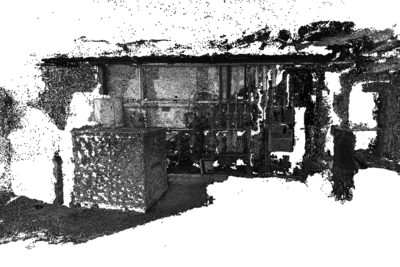}&
\includegraphics[width=0.2\textwidth]{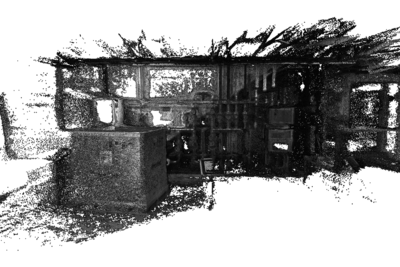}&
\includegraphics[width=0.2\textwidth]{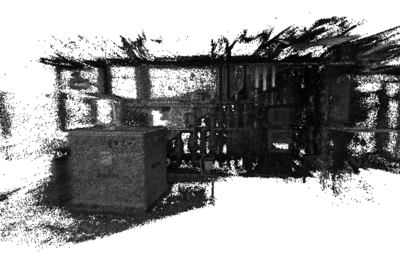}&
\includegraphics[width=0.2\textwidth]{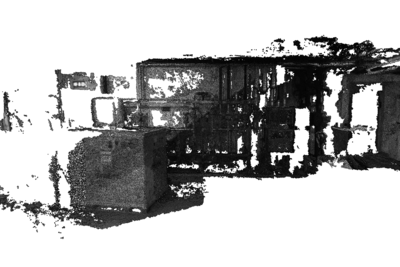}&
\includegraphics[width=0.2\textwidth]{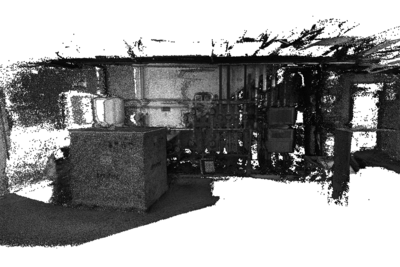}\\
\multicolumn{5}{c}{storage 2}\\
\includegraphics[width=0.2\textwidth]{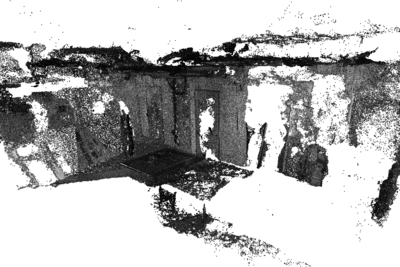}&
\includegraphics[width=0.2\textwidth]{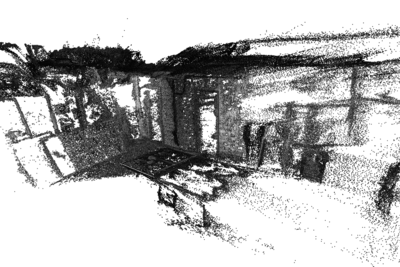}&
\includegraphics[width=0.2\textwidth]{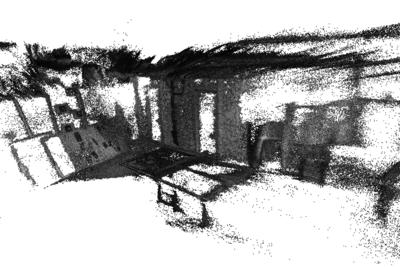}&
\includegraphics[width=0.2\textwidth]{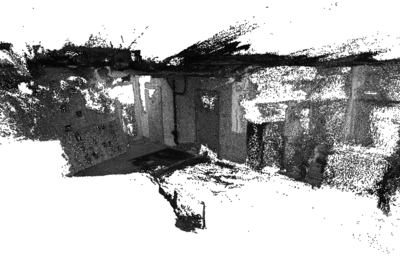}&
\includegraphics[width=0.2\textwidth]{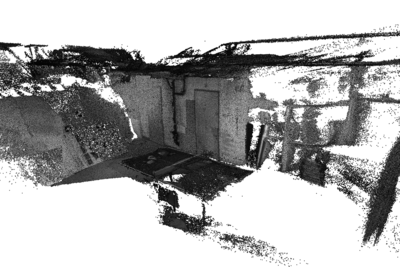}\\
\multicolumn{5}{c}{pipes}\\
\includegraphics[width=0.2\textwidth]{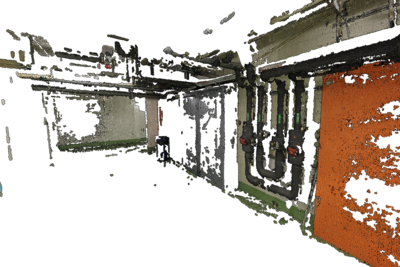}&
\includegraphics[width=0.2\textwidth]{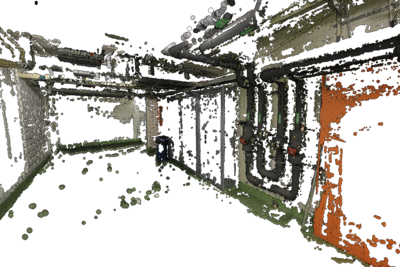}&
\includegraphics[width=0.2\textwidth]{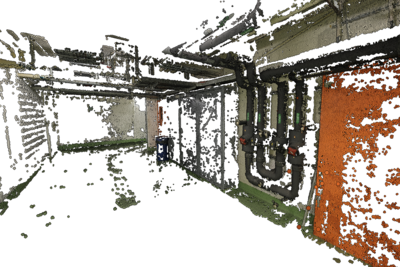}&
\includegraphics[width=0.2\textwidth]{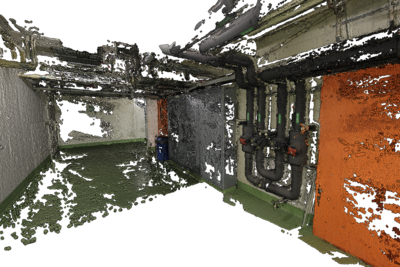}&
\includegraphics[width=0.2\textwidth]{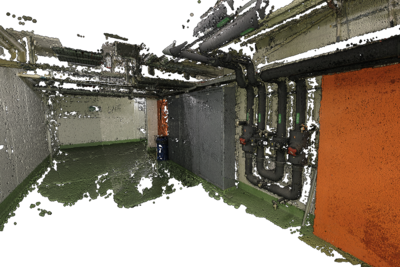}\\
\multicolumn{5}{c}{living room}\\
\includegraphics[width=0.2\textwidth]{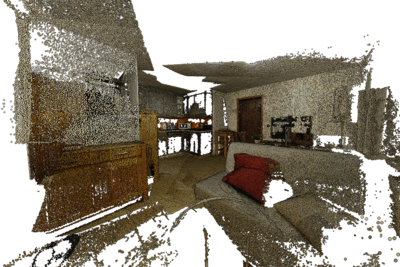}&
\includegraphics[width=0.2\textwidth]{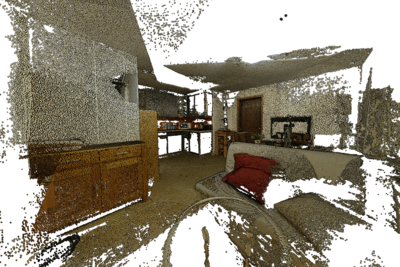}&
\includegraphics[width=0.2\textwidth]{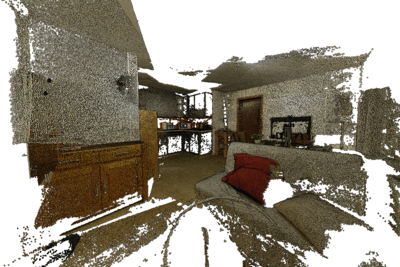}&
\includegraphics[width=0.2\textwidth]{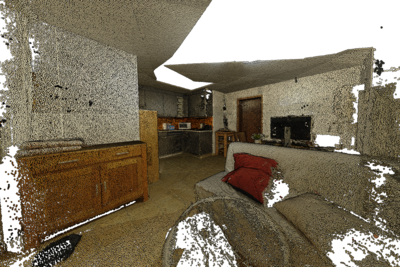}&
\includegraphics[width=0.2\textwidth]{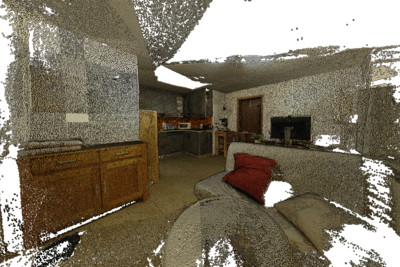}\\
\multicolumn{5}{c}{kicker}\\
\includegraphics[width=0.2\textwidth]{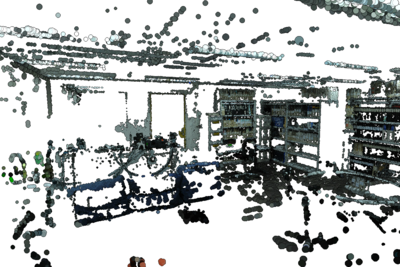}&
\includegraphics[width=0.2\textwidth]{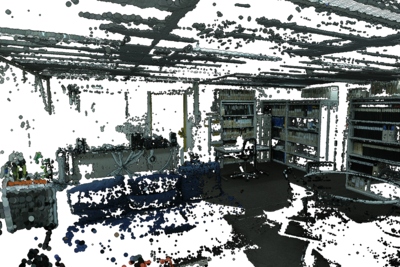}&
\includegraphics[width=0.2\textwidth]{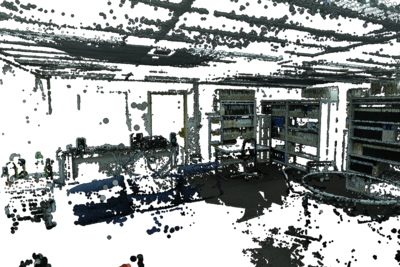}&
\includegraphics[width=0.2\textwidth]{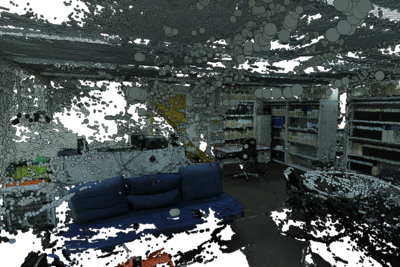}&
\includegraphics[width=0.2\textwidth]{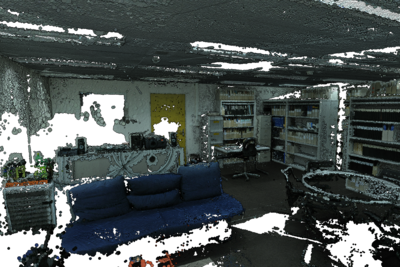}\\
LTVRE \cite{kuhn2017tv}&COLMAP \cite{schonberger2016pixelwise}&ACMH \cite{xu2018multi}&OpenMVS &TAPA-MVS (Proposed)\\
\end{tabular}
\caption{Results on ETH3D}
\label{fig:eth3D_res}
\end{center}
\end{figure*}

Table \ref{tab:ETH3D} shows the F1-scores computed with a threshold of 2 cm, which is the default values adopted for the dataset leaderboard.
TAPA-MVS, \ie, the method proposed in this paper, is ranked first according to the overall F1-score of both the Training and Test sequences.
It is worth noticing that TAPA-MVS, improves significantly the results of the baseline COLMAP framework.
The reason for such successful reconstruction has to be ascribed to the texture aware mechanism which is able to accurately reconstruct the photometrically stable areas and to recover the missing geometry where the photo-consistent measure is unreliable.
Figure \ref{fig:eth3D_res} shows the models recovered by TAPA-MVS and the top performing algorithms in some of the ETH3D sequences. 
The models reconstructed by TAPA-MVS are significantly more complete and contain less noise.

\begin{figure*}[tp!h]
\begin{tabular}{c}
 
\centering
\resizebox {0.33\textwidth} {!} {
  \begin{tikzpicture}
  \begin{axis}[
      enlargelimits=false,
      ylabel={\% pixels with L1 error $<$ 10cm},
      xlabel={textureness ($t_x$) },
       xmin=10, xmax=100,
       ymin=0, ymax=100,
      xtick={0,20,30,40,50,60,70,80,90,100},
      ytick={0,20,30,40,50,60,70,80,90,100},
      legend pos=south east,
      ymajorgrids=true,
      grid style=dashed,
  ]
  \addplot[
      color=blue,
      mark=oplus*]
  table[x index=0,y index=1,col sep=comma]
  {data/textureness10.txt};
  \addplot[
      color=red,
      mark=triangle*]
  table[x index=0,y index=2,col sep=comma]
  {data/textureness10.txt};
  \addplot[
      color=green,
      mark=square*]
  table[x index=0,y index=3,col sep=comma]
  {data/textureness10.txt};
  \legend{DeepMVS\cite{huang2018deepmvs},COLMAP\cite{schonberger2016pixelwise},TAPA-MVS(Proposed)}
  \end{axis}
  \end{tikzpicture}
}

\resizebox {0.33\textwidth} {!} {
  \begin{tikzpicture}
  \begin{axis}[
      enlargelimits=false,
      ylabel={\% pixels with L1 error $<$ 20cm},
      xlabel={\% textureness ($t_x$) },
       xmin=10, xmax=100,
       ymin=0, ymax=100,
      xtick={0,20,30,40,50,60,70,80,90,100},
      ytick={0,20,30,40,50,60,70,80,90,100},
      legend pos=south east,
      ymajorgrids=true,
      grid style=dashed,
  ]
  \addplot[
      color=blue,
      mark=oplus*]
  table[x index=0,y index=1,col sep=comma]
  {data/textureness20.txt};
  \addplot[
      color=red,
      mark=triangle*]
  table[x index=0,y index=2,col sep=comma]
  {data/textureness20.txt};
  \addplot[
      color=green,
      mark=square*]
  table[x index=0,y index=3,col sep=comma]
  {data/textureness20.txt};
  \legend{DeepMVS\cite{huang2018deepmvs},COLMAP\cite{schonberger2016pixelwise},TAPA-MVS(Proposed)}
  \end{axis}
  \end{tikzpicture}
}

\resizebox {0.33\textwidth} {!} {
  \begin{tikzpicture}
  \begin{axis}[
      enlargelimits=false,
      ylabel={\% pixels with L1 error $<$ 50cm},
      xlabel={\% textureness ($t_x$) },
       xmin=10, xmax=100,
       ymin=0, ymax=100,
      xtick={0,20,30,40,50,60,70,80,90,100},
      ytick={0,20,30,40,50,60,70,80,90,100},
      legend pos=south east,
      ymajorgrids=true,
      grid style=dashed,
  ]
  \addplot[
      color=blue,
      mark=oplus*]
  table[x index=0,y index=1,col sep=comma]
  {data/textureness50.txt};
  \addplot[
      color=red,
      mark=triangle*]
  table[x index=0,y index=2,col sep=comma]
  {data/textureness50.txt};
  \addplot[
      color=green,
      mark=square*]
  table[x index=0,y index=3,col sep=comma]
  {data/textureness50.txt};
  \legend{DeepMVS\cite{huang2018deepmvs},COLMAP\cite{schonberger2016pixelwise},TAPA-MVS(Proposed)}
  \end{axis}
  \end{tikzpicture}
}
\end{tabular}

\caption{Percentage of pixels with error $<$ 10cm, 20cm and 50cm with respect to textureness}
\label{fig:text50}
\end{figure*}
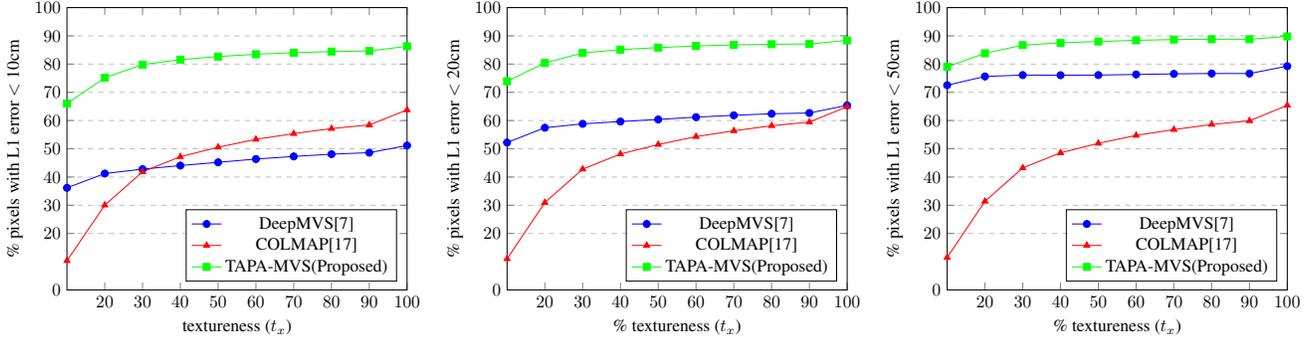

To further test the effectiveness of our method, we compared directly the accuracy of the depth map in the 13 training high-resolution sequences against the baseline COLMAP \cite{schonberger2016pixelwise} and the recent deep learning-based DeepMVS \cite{xu2018multi}.
Figure \ref{fig:depthRes} illustrates the error distribution, \ie, the percentage of pixels in the depth maps whose error is lower than a variable threshold (x-axis). TAPA-MVS clearly shows better completeness with respect to both methods, especially when considering small errors.
In Figure \ref{fig:text50} we define image regions with respect to increasing textureness values relying of the term $t_x$ described in Section \ref{sub:hypo_intr}. Given a value $v$ in the x-axis, we consider the image areas where the textureness coefficient $t_x<v$ and we plot in the three graphs the percentage of pixels in these areas with a depth error lesser than 10cm, 20 cm or 50cm. These graphs demonstrate the robustness of the proposed method against untextured regions, indeed even in low-textured areas, the percentage of pixel correctly estimated is comparable to the highly textured regions.

\begin{table*}[tp!h]
    \centering
    \setlength{\tabcolsep}{2px}
    \begin{tabular}{lcccccc}
    Method & 
    \multicolumn{3}{c}{Training sequences} &
    \multicolumn{3}{c}{Test sequences}\\
             & Overall & Low-Res & High-Res & Overall & Low-Res & High-Res \\
    TAPA-MVS (Proposed) & \textbf{71.42}   & 55.13   & \textbf{77.69}    & \textbf{73.13}   & \textbf{58.67}   & 79.15    \\ 
    OpenMVS   & 70.44   & \textbf{55.58}   & 76.15    & 72.83   & 56.18   & \textbf{79.77}    \\
    ACMH \cite{xu2018multi}    & 65.37   & 51.50   & 70.71    & 67.68   & 47.97   & 75.89    \\
    COLMAP \cite{schonberger2016pixelwise}   & 62.73   & 49.91   & 67.66    & 66.92   & 52.32   & 73.01    \\
    LTVRE \cite{kuhn2017tv}   & 59.44   & 53.25   & 61.82    & 69.57   & 53.52   & 76.25    \\
    CMPMVS \cite{jancosek2011multi}   & 47.48   & 9.53    & 62.49    & 51.72   & 7.38    & 70.19    \\
    
    \end{tabular}
    \caption{f1 scores on the ETH3D Dataset with tolerance  $\tau=$2cm (used by default for the dataset leaderboard page).}
    \label{tab:ETH3D}
\end{table*}

\begin{table*}[tp!h]
    \centering
    \setlength{\tabcolsep}{2px}
    \begin{tabular}{lcccccccccccccccccc}
    $\tau$&\multicolumn{3}{c}{COLMAP\cite{schonberger2016pixelwise}}&
    \multicolumn{3}{c}{w/o TW}&
    \multicolumn{3}{c}{w/o CS}&
    \multicolumn{3}{c}{w/o FS}&
    \multicolumn{3}{c}{w/o DR}&
    \multicolumn{3}{c}{TAPA-MVS}\\
    &C & A & F1&
    C & A & F1&
    C & A & F1&
    C & A & F1&
    C & A & F1&
    C & A & F1\\
    1   &
    38.65&\textbf{84.34}&51.99&
    32.68&74.40&44.58&
    41.72&75.30&53.18&
    41.35&75.10&52.86&
    47.78&72.13 &56.31&        
    \textbf{51.66}&75.37&\textbf{60.85}\\
    2 &
    55.13&\textbf{91.85}&67.66&
    52.57&85.70&63.08&
    64.13&85.98&72.54&
    63.69&85.77&72.26&
    64.27&83.32&71.84&
    \textbf{71.45}&85.88&\textbf{77.69}\\
    5 &
    69.91&\textbf{97.09}&80.5&
    69.31&94.08&78.62&
    81.08&93.69&86.68&
    80.84&93.58&86.51&
    78.62&92.51&84.37&
    \textbf{84.83}&94.31&\textbf{88.91}\\
    10 &
    79.47&\textbf{98.75}&87.61&
    78.10&96.91&85.64&
    88.80&96.53&92.38&
    88.61&96.45&92.22&
    86.33&95.94&90.47&
    \textbf{90.98}&96.79&\textbf{93.69}\\
    20 &
    88.24&\textbf{99.37}&93.27&
    84.93&98.34&90.53&
    93.64&98.12&95.77&
    93.61&98.05&95.72&
    91.26&97.75&94.25&
    \textbf{94.72}&98.23&\textbf{96.38}\\
    50&
    96.03&\textbf{99.70}&97.78&
    92.07&99.30&95.19&
    97.33&99.23&98.25&
    97.54&99.20&98.34&
    95.65&99.21&97.23&
    \textbf{97.60}&99.30&\textbf{98.41}\\

    \end{tabular}
    \caption{Ablation study: without Texture Weighting (TW), Coarse Superpixels (CS), Fine Superpixels (FS), Depth Refinement (DR)}
    \label{tab:ablation}
\end{table*}

\subsubsection*{Ablation study}
To assess the effectiveness of all the proposal of the paper, Table \ref{tab:ablation} shows the accuracy, completeness and F1-score of our method in the training high-resolution sequences whose ground truth is publicly available. In the table, the rows represent increasing values of the distance threshold $\tau$.
We listed the results without the Texture Weighting (TW), without using the Coarse or the Fine Superpixels (CS and FS) and finally without the Depth Refinement step (DR). We also added to the comparison the COLMAP performance \cite{schonberger2016pixelwise} which is the baseline algorithm prior to the novel steps suggested by this paper.

As expected COLMAP achieves the best accuracy at the cost of lower completeness since it produces depth estimates only in correspondence of textured regions.
The data clearly shows that all the single proposal described in the previous sections are crucial to the balance between model completeness and accuracy obtained by TAPA-MVS.
In particular, texture weighting is fundamental to avoid the framework treating the proposed hypothesis with the same importance as the old ones, no matter how much texture the image contains, this induces, in some cases, severe errors that led the optimization into local minima.
The Superpixels plane fitting steps are both relevant to obtain good guesses for untextured regions. Finally, depth refinement not only improves the completeness of the results but, by filtering out wrong estimates and replacing them with a careful neighbors interpolation at the missing estimate, it improves the accuracy as well.

\section{Conclusions and Future Works}
We presented a PatchMatch-based framework for Multi-View Stereo which is robust in correspondence of untextured regions.
By choosing a set of novel PatchMatch hypotheses, the optimization framework expands the photometrically stable depth estimates, corresponding to image edges and textured areas, to the neighboring untextured regions. We demonstrated that a modification of the cost function used by the framework to evaluate the goodness of such hypotheses is needed, in particular, by favoring the novel ones when the textureness is low.
We finally propose a depth refinement method that improves both reconstruction accuracy and completeness.

In the future, we plan to build a complete textureness-aware MVS pipeline including also a mesh reconstruction and refinement stages. In particular, we are interested in a robust meshing stage embedding piecewise planar priors, where the point clouds regions correspond to untextured areas. Moreover, we would like to define a mesh refinement method that balances regularization and data-driven optimization depending on image textureness.

{\small
\bibliographystyle{ieee}
\bibliography{biblioTotal}
}

\end{document}